\documentclass[a4paper, 10pt, conference]{ieeeconf}      
\IEEEoverridecommandlockouts                              
\usepackage{amssymb}  
\usepackage{amsmath} 
\usepackage{pdfpages}
\usepackage{multirow}
\usepackage{etoolbox}
\usepackage{threeparttable}
\usepackage{multicol}

\makeatletter
\let\NAT@parse\undefined
\patchcmd{\@makecaption} 
  {\scshape}
  {}
  {}
  {}
\makeatletter
\patchcmd{\@makecaption}
  {\\}
  {.\ }
  {}
  {}
 \newcommand{\figcaption}[1]{\def\@captype{figure}\caption{#1}}
 \newcommand{\tblcaption}[1]{\def\@captype{table}\caption{#1}}
\makeatother

\usepackage{hyperref}  

\title{\LARGE \bf
Recognition and 3D Localization of Pedestrian \\ Actions from Monocular Video
}

\author{Jun Hayakawa, Behzad Dariush
\thanks{Jun Hayakawa and Behzad Dariush are with Honda Research Institute, 70 Rio Robles, San Jose, CA 95134 USA. (Phone: +1-650-314-0400; e-mail: {jhayakawa, bdariush}@honda-ri.com).}
}

\begin{document}
\maketitle
\thispagestyle{empty}
\pagestyle{empty}

\begin{abstract}
Understanding and predicting pedestrian behavior is an important and challenging area of research for realizing safe and effective navigation strategies in automated and advanced driver assistance technologies in urban scenes. This paper focuses on monocular pedestrian action recognition and 3D localization from an egocentric view for the purpose of predicting intention and forecasting future trajectory.  A challenge in addressing this problem in urban traffic scenes is attributed to the unpredictable behavior of pedestrians, whereby actions and intentions are constantly in flux and depend on the pedestrians pose, their 3D spatial relations, and their interaction with other agents as well as with the environment.  To partially address these challenges, we consider the importance of pose toward recognition and 3D localization of pedestrian actions.
In particular, we propose an action recognition framework using a two-stream temporal relation network with inputs corresponding to the raw RGB image sequence of the tracked pedestrian as well as the pedestrian pose.  The proposed method outperforms methods using a single-stream temporal relation network based on evaluations using the JAAD public dataset. The estimated pose and associated body key-points are also used as input to a network that estimates the 3D location of the pedestrian using a unique loss function.  The evaluation of our 3D localization method on the KITTI dataset indicates the improvement of the average localization error as compared to existing state-of-the-art methods.  Finally, we conduct qualitative tests of action recognition and 3D localization on  HRI's H3D driving dataset.
\end{abstract}

\section{INTRODUCTION}
Understanding pedestrian behavior is an important and challenging problem that is critical for the deployment of automated and advanced driving assistance technologies in production vehicles. The challenges are exasperated in situations involving pedestrian interactions in dense urban environments, such as intersections. The difficulty in understanding and modeling pedestrian behavior is primarily attributed to the unpredictability of human behavior in situations where actions and intentions are constantly in flux and depend on the abrupt variations in the human pose, their 3D spatial relations, and their interaction with other agents as well as with the environment.  To arrive at a pragmatic solution, we emphasize the importance of human pose estimation for recognition and 3D localization of actions.

The core technical components which are essential for understanding pedestrian behavior include recognition and 3D localization of actions and prediction of intention and future trajectory in traffic scenes.  While human action recognition and localization problems involve estimation of the current state (behavior and location), future trajectory prediction involves estimation of the future path based on the present and past observations.  Therefore, real-time analysis of the current state provides an informative prior in forecasting future trajectories for use in decision making and path planning strategies.   Furthermore, it is important to localize and forecast the pedestrian motion in 3D, in order to properly assess risk and develop countermeasures using various driving assistance methodologies.

Recent papers \cite{StarNet, alahiSocialLSTMHuman2016,guptaSocialGANSocially2018,choiLookingRelationsFuture, yaoEgocentricVisionbasedFuture2019, titanHRI} related to trajectory prediction of road agents localize the 2D position of the agents in the image and predict future trajectories using GRU, LSTM, and GAN.  While these methods have achieved good performance on several public datasets, forecasting future 3D-trajectory of pedestrians in urban driving scenes remains a challenge due to: (a) inaccuracies in 3D pedestrian localization from images, (b) uncertainty of the human intention, and (c) complex interactions between traffic participants and the environment. 

To address interactions between agents, influential work such as Social LSTM \cite{alahiSocialLSTMHuman2016} and Social GAN \cite{guptaSocialGANSocially2018} explicitly modeled the interactions between traffic participants to forecast future trajectories from a third-person view.  More recently, future vehicle trajectory forecast from an egocentric view was introduced where object location and scale level observations were used \cite{yaoEgocentricVisionbasedFuture2019}. The work by \cite{choiLookingRelationsFuture} proposed a method for the relation-aware framework for future trajectory forecast. The most recent approach \cite{titanHRI}, introduced a trajectory inference approach using Target Action priors Network to predict the future trajectory of scene agents.  
While the aforementioned methods have made significant advances in trajectory forecast in 2D image coordinates, more accurate and detailed representations of the pedestrian state, including pedestrian action and 3D localization, are needed as inputs to improve accuracy and robustness in driving scenes.

\begin{figure*}[ht]
    \centering
    \includegraphics[scale=0.6]{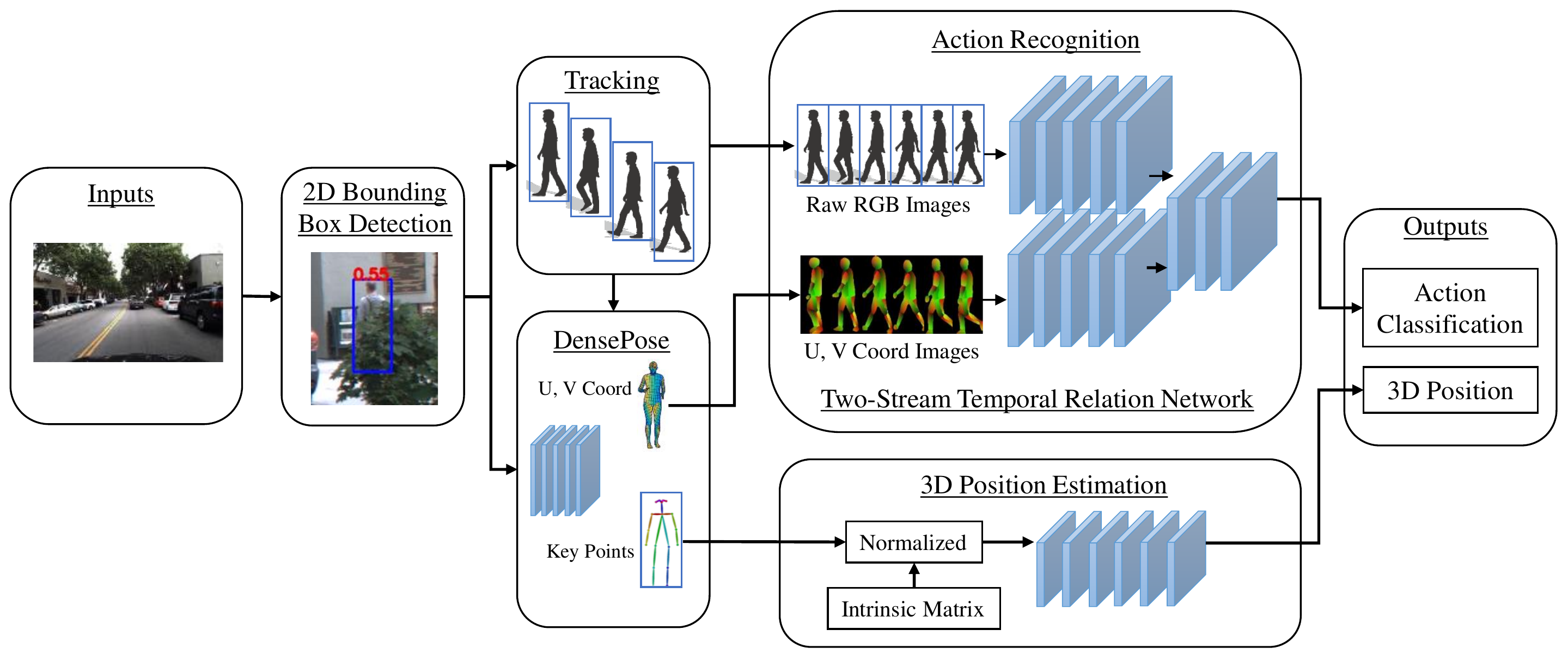}
\caption{The overview of our approach for action recognition and 3D localization. The two-stream temporal relation network classifies pedestrian action from eight sequential images of raw RGB and DensePose (U, V coordinate). 3D localization network estimates 3D position in ego-centered coordinate system from human 2D pose using a new loss function of Johnson SU distribution.}
\label{fig:overview}
\end{figure*}

This paper focuses on accurate action recognition and 3D localization of pedestrians in street scenes. We present a two-stream temporal relation network using images of raw RGB and human pose (using DensePose \cite{guler2018densepose}) as network inputs to recognize pedestrian actions based on an existing temporal relation network \cite{zhouTemporalRelationalReasoning2018}. Temporal relation networks can learn and reason about temporal dependencies between video frames at multiple time scales. Recent research \cite{yanSpatialTemporalGraph2018} has shown a significant improvement of action recognition based on key points of human joints and the spatial relation between those key points.  While the key point information has been helpful to improve performance of action recognition in close proximity using only raw RGB images, they are less informative for a representation of pedestrian state in traffic scenes where the resolution of the human is often coarse. DensePose \cite{guler2018densepose} is a state-of-the-art method of pose estimation that also outputs more extensive information about pedestrians. 

The contributions of this paper are two-fold. First, we use images of raw RGB and DensePose as inputs to a two-stream temporal relation network in order to improve the accuracy of action recognition. We demonstrate the efficacy of our approach through comparisons against single-stream temporal relation networks using the JAAD dataset \cite{kotserubaJointAttentionAutonomous2017}. Second, we introduce a new loss function to the existing 3D localization approach, MonoLoco \cite{bertoniMonoLocoMonocular3D2019}. Our loss function encodes pedestrian key-point information which considers the asymmetric distribution of the distance error converted from the key points on the image plane. We evaluate our 3D localization method on the KITTI dataset \cite{Geiger2013IJRR}. Finally, we show qualitative results of experiments on action recognition and 3D localization on HRI's H3D driving dataset \cite{360LiDARTracking_ICRA_2019}. 

\section{Related Work}

\textbf{2D Bounding Box Detection and Tracking.} Novel deep learning methods improve the accuracy of 2D bounding box detection. 2D object detection algorithms such as YOLO V3 \cite{redmonYOLOv3IncrementalImprovement2018}, SSD \cite{liuSSDSingleShot2016}, and Mask R-CNN \cite{heMaskRCNN2018} have achieved high performance with fast and accurate 2D object detection using a monocular camera. CBNet \cite{liuCBNetNovelComposite2019} realizes significant improvements in detection accuracy using ResNeXt as a backbone network for feature extraction. CSP \cite{liu2018high} can detect a 2D bounding box of pedestrians, including occluded human body parts. Pedestrians are often occluded by other traffic participants and road objects. Therefore, the whole bounding box, including occluded human body part, helps to track pedestrians based on its bounding box size and position.

\textbf{Pose Estimation.}
Recent results \cite{wei2016cpm, toshevDeepPoseHumanPose2014, kreiss2019pifpaf} analyze joints of a human body and output key points on the image coordinates. Moreover, 3D pose estimation from a monocular image \cite{yang3DHumanPose2018} proposes an adversarial learning framework that can estimate the 3D human pose structures learned from the fully annotated dataset with only 2D pose annotations. The state-of-the-art pose estimation is DensePose \cite{guler2018densepose}. DensePose outputs 3D surface-based representation in surface coordinates of (SMPL) model \cite{loperSMPLSkinnedMultiperson2015} as well as key points of pedestrians. Therefore, DensePose can represent the human body in much more detail than simply providing key points.
 
\textbf{Action Recognition.}
There are two main approaches for action recognition, image-based \cite{wangTemporalSegmentNetworks2016, zhouTemporalRelationalReasoning2018,rasouli2017they}, and skeleton-based \cite{liActionalStructuralGraphConvolutional, siAttentionEnhancedGraph2019, yanSpatialTemporalGraph2018}. Both methods capture spatial and temporal information from sequential input images or key points of pedestrians. Image-based methods have an advantage in that they can detect pedestrian actions by exploiting contextual information of the road environments.  Skeleton-based approaches can classify actions from human pose features and have a benefit in terms of computation speed. 

\textbf{3D Localization.}
3D localization from a monocular camera image sequence is a challenging research topic. Mono3D \cite{chenMonocular3DObject2016} assumes objects are on the ground plane in order to regress the 3D location and 3D bounding box of the object.  This assumption creates difficulties when pedestrians are on different planes, such as sidewalks or on slopes. MonoDepth \cite{godardUnsupervisedMonocularDepth2016} estimates depth from a single monocular image. Stereo-based research \cite{3dop} outperforms methods using a single monocular camera in terms of accuracy. However, stereo systems are more costly and introduce additional complexities such as requiring high precision calibration. MonoLoco \cite{bertoniMonoLocoMonocular3D2019} is a computationally efficient approach using a lightweight network that predicts 3D locations from 2D human pose, taking into account the uncertainty in depth.


\begin{figure}[t]
    \centering
    \includegraphics[scale=0.4]{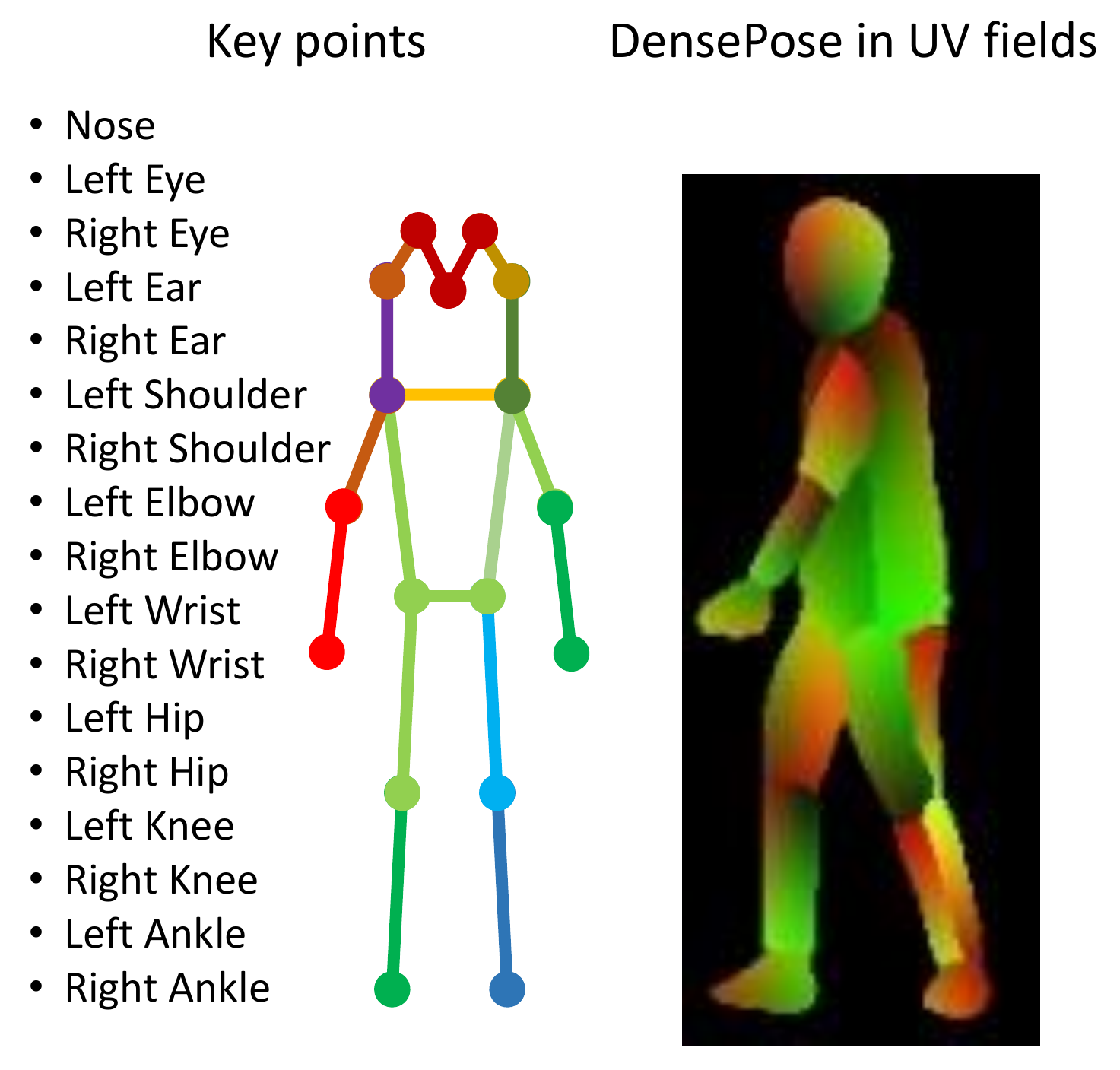}
\caption{Left: 17 Key points representing the human joints in the image plane. Right: DensePose image in UV fields, resulting in 24 parts of the human body surface based on Skinned Multi-Person Linear (SMPL) model.  We use the DensePose image for action recognition.}
\label{fig:densepose}
\end{figure}

\section{Approach for Action Recognition and 3D Localization}

In this section, we present our method for pedestrian action recognition and localization. Fig. \ref{fig:overview} shows an overview of our approach. We estimate pedestrian action from raw RGB and DensePose images using a two-stream temporal relation network. Simultaneously, we predict 3D locations from key points of pedestrians using our proposed loss function.

\subsection{2D Bounding Box Detection and Tracking}
2D bounding box detection of pedestrians is the first step in the proposed method. We use CSP \cite{liu2018high} to detect the 2D bounding box because  CSP generates the entire bounding boxes even if the pedestrian is occluded. We use the pre-trained model from the City Person dataset \cite{zhangCityPersonsDiverseDataset2017}. Subsequently, the pedestrian bounding box is tracked between successive frames using DeepSort \cite{Wojke2017simple}, which is computationally more efficient than feature-based tracking methods. DeepSort also has a high affinity with CSP because CSP is more robust to occluded pedestrians.
The DeepSort CNN model has been trained on a large-scale person re-identification dataset \cite{zheng2016mars}.

\subsection{Pose Estimation}
We use DensePose \cite{guler2018densepose} to estimate the pose and key point locations on the body. DensePose has a cross-cascading architecture that includes the DensePose network to generate the body pose and auxiliary networks to generate key points and masks.  In particular, the DensePose network generates a colored image representing 24 parts of the human body surface in the U, V coordinate. The auxiliary networks simultaneously generate key points. Fig.\ref{fig:densepose} shows image samples of pose estimation results. As shown in Fig.\ref{fig:densepose}, the DensePose image has rich information compared with Key points of a pedestrian. We use a pre-trained model from DensePose-COCO dataset that is introduced in \cite{guler2018densepose}.

\subsection{Pedestrian Action Recognition}
\begin{figure}[t]
    \centering
    \includegraphics[scale=0.54]{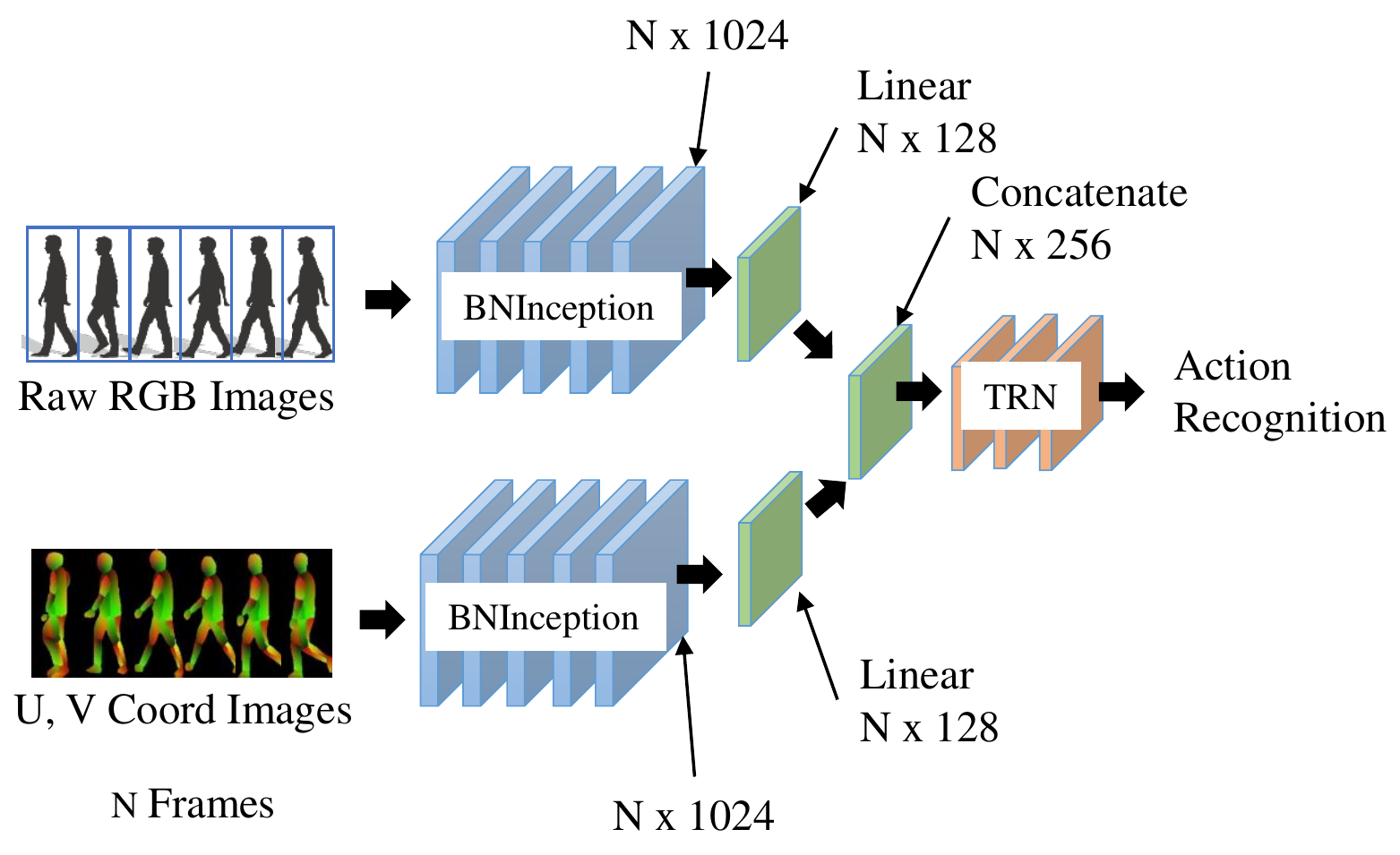}
\caption{Our network structure for action recognition. We add a linear layer after the final layer of Batch Normalized Inception outputs$\{N, 1024\}$ and concatenate the outputs of each stream $f_i=\{N, 256\}$.}
\label{fig:network}
\end{figure}

We propose a two-stream temporal relation network using images of raw RGB and DensePose in the U, V coordinate based on existing research \cite{zhouTemporalRelationalReasoning2018}. The unique feature of our approach is to use the DensePose images as one of the inputs to the two-stream temporal relational network because DensePose can represent the detailed human pose such as the direction and size of each body part compared to joint key points of human joints. We crop images of the raw RGB and DensePose according to the size of the 2D bounding box. We adopt Inception with Batch Normalization pre-trained on ImageNet as a base network model for feature extraction for both raw RGB and DensePose images. We add a linear layer at the output of the final layer of Batch Normalized Inception outputs$\{N, 1024\}$ and concatenate the outputs  $f_i=\{N, 256\}$, where $i$ is $i^{th}$ frame. Moreover, we define $V=\{f_1, f_2,...f_N\}$ as input for the temporal relation network. Our network structure is shown in Fig.\ref{fig:network}.

\begin{equation}
\label{eq:action1}
T_N(V) = h'_\phi \left( \sum_{i<j...<N}g'_\theta (f_i,f_j...,f_N) \right)
\end{equation}
where N is the total frames required to capture a temporal relation. Furthermore, we use a multi-scale temporal relation network \cite{zhouTemporalRelationalReasoning2018} to understand temporal relations at multiple time scales. 
\begin{equation}
\label{eq:action2}
MT_N(V) = T_2(V) + T_3(V) ... + T_N(V)
\end{equation}
Each $T_d$ represents the temporal relation between $d$ ordered frames.

\subsection{3D Localization of pedestrians}
The aim of the 3D localization module is to estimate the 3D position of each pedestrian with respect to the ego-vehicle from a single monocular camera. We assume aleatoric uncertainty captured by the probability distribution based on MonoLoco  \cite{bertoniMonoLocoMonocular3D2019}. MonoLoco uses symmetrically distributed loss functions such as Gaussian and Laplace loss. However, a pixel error of the key points on the image plane affects the distance estimation accuracy in a way that depends on the distance to the pedestrian. For example, distance estimation accuracy for pedestrians at proximity is less affected by the pixel error of the key points, while the accuracy for distant pedestrians are heavily affected. The distance estimation error as a result of the pixel error of key points is remarkably large for distant objects. Therefore, the distance errors are not distributed symmetrically, but rather asymmetrically. In our approach, we define an asymmetric distribution that is negative log-likelihood of Johnson SU loss function for the representation of aleatoric uncertainty. Johnson SU distribution can represent symmetric or asymmetric distribution using four parameters. Johnson SU distribution is significantly flexible and robust to address the distance error. Inputs of our method are key points of pedestrians on the image plane, and the neural network regresses the distance from the ego-vehicle to each pedestrian.  Our proposed Johnson SU loss function is described as: 

\begin{equation}
\label{eq:localization1}
\begin{split}
L_J(x|\gamma , \delta , \lambda , \xi) = \\
&(\gamma + {\delta sinh^{-1}z})^{2} -log(\delta)  \\
&+log \left( \frac{1}{\lambda \sqrt{2\pi} \sqrt{z^{2} + 1}} \right)
\end{split}
\end{equation}
where $z$ is as:

\begin{equation}
\label{eq:localization2}
z = \frac{x-\xi}{\lambda}
\end{equation}
where $x$ is the ground truth of the distance. $\{\gamma , \delta , \lambda , \xi\}$ are parameters learned by the model ,and $\xi$ is the estimated distance. The main advantage of Johnson SU loss function is to improve 3D localization accuracy for distant objects.

Although MonoLoco \cite{bertoniMonoLocoMonocular3D2019} uses PifPaf \cite{kreiss2019pifpaf} to detect key points of pedestrians, our pipeline uses DensePose \cite{guler2018densepose} in order to simplify the algorithm and reduce the computational complexity.

\begin{table}[t]
\caption{Comparison result of action recognition using Temporal Relation Network.}
\label{table_action_TRN_result}
\begin{center}
\begin{tabular}{c|c c c|c}
 & TRN RGB & TRN Flow & TRN DensePose & {\shortstack[1]{Accuracy\\(\%)}} \\
\hline
\hline
\multirow{3}{*}{\shortstack[1]{Single\\ Stream}} &  \checkmark & & & 36.8\\
& & \checkmark & & 35.7\\
& & & \checkmark & 30.6\\
\hline
\multirow{3}{*}{\shortstack[1]{Two\\ Stream}} &  \checkmark & \checkmark & & 40.0\\
& \checkmark & & \checkmark & \textbf{42.2}\\
\end{tabular}
\end{center}
\end{table}

\section{Evaluation}
In this section, we provide a qualitative and quantitative evaluation of our action recognition and 3D localization algorithm shown in Fig. \ref{fig:overview} using publicly available datasets.  
\subsection{Action Recognition} 
We assess our two-stream action recognition algorithm with images of raw RGB and DensePose using JAAD dataset \cite{kotserubaJointAttentionAutonomous2017}. The JAAD dataset provides timestamped behavior labels and 2D bounding boxes of pedestrians. Moreover, the JAAD dataset has demographic attributes for each pedestrian (e.g., gender, age, etc.). We selected the following eight pedestrian action labels from the JAAD dataset:

\begin{multicols}{2}
\begin{itemize}
\item Walking
\item Nodding
\item Looking at Ego-Vehicle
\item Crossing Streets
\item Clearing Path
\item Speed Up
\item Speed Down
\item Making Hand Gesture
\end{itemize}
\end{multicols}

The total number of prepared frames is 47K: 42K frames are for training, and 5K frames are for evaluation. As defined in Eq (\ref{eq:action1}) and Eq (\ref{eq:action2}), we use $N=8$ total frames to capture a temporal relation in this experiment. We created DensePose images using the pre-trained DensePose model from DensePose-COCO dataset. We compared our two-stream temporal relation network using images of raw RGB and DensePose against a two-stream using raw RGB and optical flow.  We also evaluated the accuracy using a single-stream temporal relation network method.  The optical flow was calculated using PWC-net \cite{Sun2018PWC-Net}. The comparative results are shown in TABLE \ref{table_action_TRN_result}.  As shown, our two-stream temporal relation network has the best performance.

\textbf{Ablation Study.} We evaluated our two-stream action recognition method using images of raw RGB and DensePose in a temporal segment network (TSN) \cite{wangTemporalSegmentNetworks2016} instead of the temporal relational network (TRN). This ablation study was done to confirm that DensePose images contribute to improvements in action recognition accuracy. The result using the temporal segment network is shown in TABLE \ref{table_action_TSN_result}. The experiment with images of raw RGB and DensePose performs better than other methods. 

\begin{table}[t]
\caption{Ablation study result of action recognition using Temporal Segment Network.}
\label{table_action_TSN_result}
\begin{center}
\begin{threeparttable}
\begin{tabular}{c|c c c|c}
 & TSN RGB & TSN Flow & TSN DensePose & {\shortstack[1]{Accuracy\\(\%)}} \\
\hline
\hline
\multirow{3}{*}{\shortstack[1]{Single\\ Stream}} &  \checkmark & & & 35.8\\
& & \checkmark & & 34.7\\
& & & \checkmark & 34.7\\
\hline
\multirow{3}{*}{\shortstack[1]{Two\\ Stream}} &  \checkmark & \checkmark & & 38.9\\
& \checkmark & & \checkmark & \textbf{40.0}\\
\end{tabular}
\end{threeparttable}
\end{center}
\end{table}

\subsection{3D localization}
We evaluated our 3D localization algorithm on the KITTI dataset by analyzing the average localization error(ALE) with respect to the ground-truth distance. We compared the results against the existing state-of-the-art methods, Mono3D \cite{chenMonocular3DObject2016}, MonoDepth \cite{godardUnsupervisedMonocularDepth2016}, Geometric \cite{bertoniMonoLocoMonocular3D2019}, and MonoLoco-baseline \cite{bertoniMonoLocoMonocular3D2019}. 

\begin{table*}[tb]
\caption{Comparison result of 3D localization.}
\label{table_localization_result}
\begin{center}
\begin{tabular}{c|c|c|c|c|c|c|c|c}
\hline
& Ground Truth Distance [m] & 3 & 8 & 12.5 & 17.5 & 22.5 & 27.5 & 35\\
\hline
\hline
&  &  \multicolumn{7}{|c}{Average Localization Error [m]}\\
\hline
\hline
Stereo & 3DOP \cite{3dop} & 0.69 & 0.40	& 0.71 & 0.93 & 1.29 & 1.69 & 4.03\\
\hline
\multirow{4}{*}{Monocular} & Mono3D \cite{chenMonocular3DObject2016} & 1.75 & 1.12 & 2.47 & 3.69 & 3.49 & 4.98 & 4.23\\
& Geometric \cite{bertoniMonoLocoMonocular3D2019} & 0.96 & 1.15 & 1.18 & 1.41 & 1.76 & 2.19 & 2.98\\
& MonoLoco-baseline \cite{bertoniMonoLocoMonocular3D2019} & 0.57 & 0.65 & 0.97 & \textbf{1.12} & 1.62 & 1.41 & 3.02\\
& Ours & \textbf{0.49} & \textbf{0.63} & \textbf{0.96} & 1.16 & \textbf{1.55} & \textbf{1.35} & \textbf{2.92}\\

\end{tabular}
\end{center}
\end{table*}

\textbf{Mono3D.} Mono3D is a 3D object detector using a single monocular camera. Mono3D assumes that objects are lying on the ground plane.

\textbf{MonoDepth.}  MonoDepth provides a single image depth estimation. We calculated the depth corresponding to the key points from DensePose and converted to the distance using the normalized image coordinate.

\textbf{3DOP.} 3DOP uses a stereo image for pedestrian detection. We referred to their publicly available results.

\textbf{MonoLoco-baseline} Original MonoLoco uses PifPaf \cite{kreiss2019pifpaf} for key point detection. We replace PifPaf with DensePose in our approach.

The comparison results are shown in TABLE \ref{table_localization_result}. Our method achieves a larger improvement at the near ($3m$) and far($22.5m - 35m$) pedestrian distances. On the other hand, our results at the middle-distance($8m-17.5m$) are similar to MonoLoco. We conjecture that the closer pedestrians are to the ego-vehicle, the more truncated. Therefore, a distance error distribution is asymmetric for close pedestrians. In addition, the distance error distribution becomes more asymmetric for more distant pedestrians. We hypothesize that the Johnson SU loss function can represent distance error precisely, especially for the near and distant pedestrians.

\begin{figure*}[tb]
    \centering
    \includegraphics[scale=0.63]{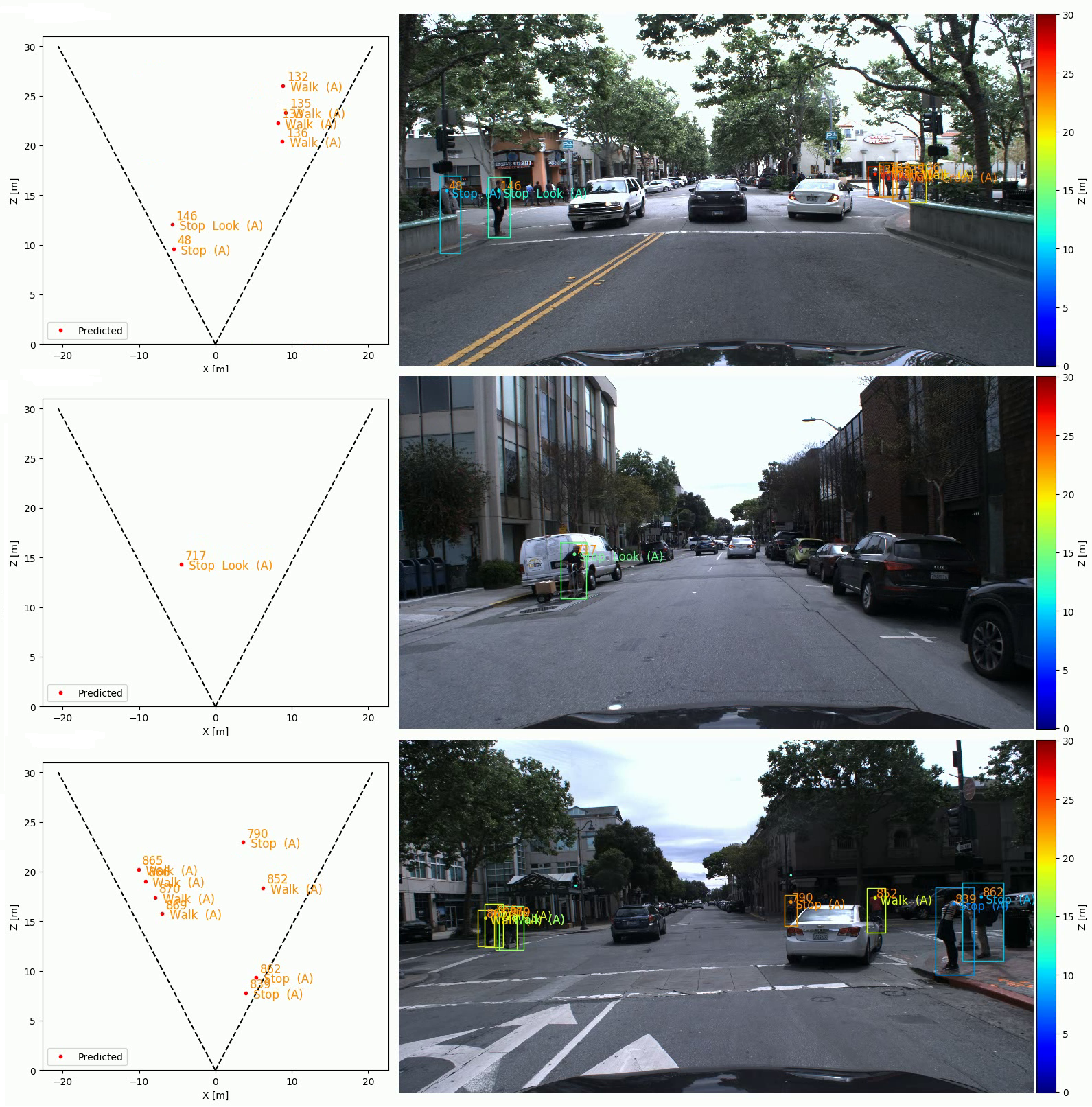}
\caption{Qualitative Evaluation with HRI's H3D Driving Dataset \cite{360LiDARTracking_ICRA_2019}. The action recognition results are shown next to the bounding boxes and localization dots.}
\label{fig:qualitative}
\end{figure*}

\subsection{Qualitative Evaluation}
We conduct qualitative tests on HRI's H3D driving dataset \cite{360LiDARTracking_ICRA_2019}. The results are shown in Fig. \ref{fig:qualitative}. Qualitatively, our action recognition and 3D localization method can recognize and localize actions for pedestrians reasonably well, even if they are occluded. Our approach has a significant advantage for partially occluded pedestrians because Dense Pose and key points can be calculated for the visible part of each pedestrian.

\section{CONCLUSIONS}
We introduced a monocular pedestrian action recognition and 3D localization approach from an egocentric view using raw RGB images and pose as inputs. The action recognition module makes use of a two-stream temporal relation network with inputs corresponding to the tracked pedestrian in the image as well as DensePose outputs. The proposed method outperforms single-stream temporal relation network methods on evaluations using the JAAD dataset. We also extended and made improvements to the method of MonoLoco for estimating the 3D locations of pedestrians by using a unique loss function of Johnosn SU distribution.  Evaluations on the KITTI dataset indicated that our method improves the average localization error as compared to existing state-of-the-art methods.  In future work, we plan to use these results to make predictions on the intention of agents in the scene and ultimately forecast their future trajectories.



\bibliographystyle{unsrt}
\bibliography{citation}

\end{document}